\documentclass[sigconf]{acmart}

\usepackage{hyperref}
\usepackage{graphicx}
\usepackage{subcaption}
\usepackage{mdframed} 
\usepackage{algorithm} 
\usepackage{algpseudocode} 
\usepackage{listings}
\usepackage{xcolor}
\usepackage{tikz}
\usepackage{balance}
\usepackage{float}

\usetikzlibrary{shapes, arrows}

\definecolor{codegreen}{rgb}{0,0.6,0}
\definecolor{codegray}{rgb}{0.5,0.5,0.5}
\definecolor{codepurple}{rgb}{0.58,0,0.82}
\definecolor{backcolour}{rgb}{0.95,0.95,0.92}

\lstdefinestyle{promptstyle}{
    backgroundcolor=\color{backcolour},   
    commentstyle=\color{codegreen},
    keywordstyle=\color{magenta},
    numberstyle=\tiny\color{codegray},
    stringstyle=\color{codepurple},
    basicstyle=\footnotesize\ttfamily,
    breakatwhitespace=false,         
    breaklines=true,                 
    captionpos=b,                    
    keepspaces=true,                 
    numbers=left,                    
    numbersep=5pt,                  
    showspaces=false,                
    showstringspaces=false,
    showtabs=false,                  
    tabsize=2
}

\AtBeginDocument{%
  \providecommand\BibTeX{{%
    Bib\TeX}}}

\copyrightyear{2024}
\acmYear{2024}
\setcopyright{acmlicensed}\acmConference[L@S '24]{Proceedings of the Eleventh ACM Conference on Learning @ Scale}{July 18--20, 2024}{Atlanta, GA, USA}
\acmBooktitle{Proceedings of the Eleventh ACM Conference on Learning @ Scale (L@S '24), July 18--20, 2024, Atlanta, GA, USA}
\acmDOI{10.1145/3657604.3662027}
\acmISBN{979-8-4007-0633-2/24/07}
\acmSubmissionID{lsfp0754}

\begin{document}

\newcommand{\moussa}[1]{\textcolor{red}{moussa: #1}}






\title{Handwritten Code Recognition for Pen-and-Paper CS Education}





\author{Md Sazzad Islam}
\orcid{0009-0002-6512-7419} 
\authornote{Equal contribution; more junior author listed first.}
\email{sazzad14@stanford.edu}
\affiliation{%
  \institution{Stanford University}
  \city{Palo Alto}
  \state{CA}
  \country{USA}
}

\author{Moussa Koulako Bala Doumbouya}
\orcid{0000-0002-5655-4489} 
\authornotemark[1] 
\email{moussa@stanford.edu}
\affiliation{%
  \institution{GNCode}
  \city{Fria}
  \country{Guinea}
}
\affiliation{%
  \institution{Stanford University}
  \city{Palo Alto}
  \state{CA}
  \country{USA}
}


\author{Christopher D. Manning}
\orcid{0000-0001-6155-649X} 
\email{manning@stanford.edu}
\affiliation{%
  \institution{Stanford University}
  \city{Palo Alto}
  \state{CA}
  \country{USA}
}

\author{Chris Piech}
\orcid{0000-0001-5140-0467} 
\email{piech@cs.stanford.edu}
\affiliation{%
  \institution{Stanford University}
  \city{Palo Alto}
  \state{CA}
  \country{USA}
}
\renewcommand{\shortauthors}{Md Sazzad Islam, Moussa Koulako Bala Doumbouya, Christopher D. Manning, \& Chris Piech}

\begin{abstract}
    Teaching Computer Science (CS) by having students write programs by hand on paper has key pedagogical advantages: It allows focused learning and requires careful thinking compared to the use of Integrated Development Environments (IDEs) with intelligent support tools or "just trying things out".
The familiar environment of pens and paper also lessens the cognitive load of students with no prior experience with computers, for whom the mere basic usage of computers can be intimidating.
Finally, this teaching approach opens learning opportunities to students with limited access to computers. 
However, a key obstacle is the current lack of teaching methods and support software for working with and running handwritten programs. 
Optical character recognition (OCR) of handwritten code is challenging: Minor OCR errors, perhaps due to varied handwriting styles, easily make code not run, and recognizing indentation is crucial for languages like Python but is difficult to do due to inconsistent horizontal spacing in handwriting.
Our approach integrates two innovative methods. The first combines OCR with an indentation recognition module and a language model designed for post-OCR error correction without introducing hallucinations. This method, to our knowledge, surpasses all existing systems in handwritten code recognition. It reduces error from 30\% in the state of the art to 5\% with minimal hallucination of logical fixes to student programs. The second method leverages a multimodal language model to recognize handwritten programs in an end-to-end fashion. We hope this contribution can stimulate further pedagogical research and contribute to the goal of making CS education universally accessible. We release a dataset of handwritten programs and code to support future research \footnote{\href{https://github.com/mdoumbouya/codeocr}{https://github.com/mdoumbouya/codeocr}}.

\end{abstract}




\received{16 February 2024}
\received[Accepted]{8 April 2024}


\maketitle

\section{Introduction}
Handwriting-based computer programming teaching tools have the potential to increase the accessibility and effectiveness of elementary CS education programs. However, to date, there aren't any usable software tools to support this pedagogy.

Handwriting-based elementary CS curricula, in contrast to those based on the use of integrated development environments (IDE), are advantageous in several ways. 
First, they reduce children's exposure to screens, which has been linked to adverse outcomes
\cite{stiglic2019effects, manwell2022digital}.
Second, they offer a more familiar and less intimidating learning environment for students with no prior experience with computers and teachers with limited experience in computer science \cite{celepkolu2020upper}. 
Third, they allow students to become more intimately familiar with code syntax as handwriting is more optimal for learning \cite{smoker2009comparing}, particularly for young children \cite{ose2020importance}.
Finally, they are less costly to implement and accessible to most schools and students.

However, to be effective, such curricula require a handwritten code recognition and execution tool that allows the student to quickly execute and test their handwritten programs. 
Automated optical handwritten code recognition is challenging because of variations in individual student handwriting features, which include character shapes, line slants, and vertical spacings \cite{srihari2002individuality}. For programming languages in which indentations are lexemes, such as Python, variations in the horizontal width of indentation units pose additional recognition challenges.

In this work, we compare various approaches to addressing the above challenges. On the one hand, we employ modular systems that include distinct modules for optical character recognition, discrete indentation level recognition, and language-model-based post-correction. On the other hand, we employ a multimodal language model that performs handwritten code recognition in an end-to-end fashion. We discuss the successes and limitations of the approaches we tried, including tradeoffs between recognition fidelity and language model-induced hallucinations.

The main contributions of this paper are:
\begin{enumerate}
    \item Provide two first-of-their-kind public benchmark datasets for handwritten student-code recognition and a methodology for measuring the correctness of a given recognition method.
    \item Contribute two novel methods of indentation recognition for handwritten Python code.
    \item Provide a novel methodology of incorporating OCR for handwritten student code with LLMs to achieve a 5\% error rate, an improvement from the state-of-the-art 30\% error.
    
\end{enumerate}

\subsection{Related Work}

\begin{figure*}[ht]
    \centering
    \includegraphics[width=\textwidth, trim=0 19cm 0 0, clip]{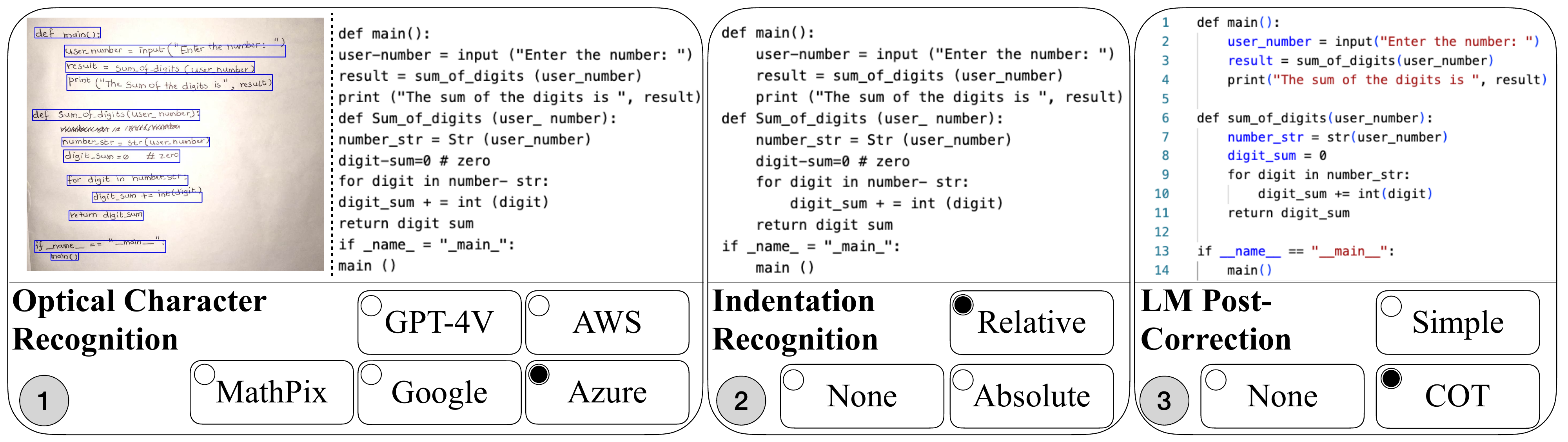}

    \caption{Processing a student's handwritten program. 
(1) The OCR module produces bounding boxes and noisy transcriptions for each line of code.
(2) High-fidelity reconstruction of the student's intended discrete indentation levels.
(3) Post-correction using chain-of-thought prompting of a language model. 
A key challenge is to reconstruct the student's work by correcting transcription errors (e.g. 
missing underscores on lines 11 and 13, dash on lines 2, 8, and 9, capitalization and inserted space on line 6) without introducing artifacts (e.g. removed comment on line 8).}
    \label{fig:methods}
\end{figure*}

Over the years, several attempts have been made to address the challenges of making computer science education more accessible. Additionally, recognizing and digitizing handwritten texts, particularly code, has been a subject of study with its unique challenges.

\paragraph{Computer Science Education without Computers}

Several initiatives have explored the teaching of computer science principles without the need for physical computers. Approaches such as the popular CS Unplugged \cite{bell2002computer, bell2018cs} curriculum use off-computer activities to teach core concepts \cite{kong2020teacher, sun2021improving}. This is especially the case for young children \cite{relkin2021learning}. However, the lack of an execution environment remains a challenge.

\paragraph{Handwriting Recognition and Digitization}

Optical Character Recognition (OCR) systems have been worked on since as early as 1912 when Emanuel Goldberg presented a machine that could convert characters into telegraph code \cite{herbert1982history} and 1913 when Dr. Edmund Fournier d'Albe created the Optophone \cite{d1914type}. OCR became a well-studied area for computing in the 1980s \cite{nagy1992frontiers, mori1992historical} where much of the focus was on identifying computer-printed characters. OCR for handwritten characters is an especially difficult OCR task \cite{memon2020handwritten, arica2002optical}. The introduction of the famous MNIST dataset \cite{cohen2017emnist} in 2013 became a popular application of deep learning algorithms which in turn led to a sharp rise in accuracy when recognizing single handwritten characters \cite{keysers2016multi}. Since then OCR of handwriting has been applied to multiple languages \cite{kumar2011review} as well as ancient scripts \cite{narang2020ancient}. However, until 2020 long form, OCR  remained relatively inaccurate, unless an algorithm was given a large sample of a particular person's handwriting. 
Integrating Large Language Models (LLMs) \cite{achiam2023gpt, bommasani2021opportunities} with OCR for post-correction has created excitement and new advances in accuracy for OCR \cite{liu2023hidden} especially using large multi-modal models such as GPT-4V \cite{hu2023bliva, shi2023exploring}. These models have been recently developed and to the best of our knowledge have not been incorporated into state-of-the-art available OCR tools. While they are promising for OCR in general, the use of LLMs needs intentional study in the context of student code, as LLMs may start solving student coding tasks while performing OCR.

\paragraph{OCR for Student Code}

While there has been a lot of work on OCR in general, and a small amount of work on OCR applied to recognizing computer printed code \cite{malkadi2020study} there is a surprising lack of work on handwritten code, let alone student handwritten code. The problem was first introduced in 2012 in the paper CodeRunner \cite{du2012code}. Perhaps because of a lack of a public dataset, little progress has been made. A project, Handwritten Code Scanner from 2021  could not progress because of how surprisingly hard it was to get standard handwritten OCR packages to work for student code \cite{handwritten_code_scanner}.

\section{The Student-Code OCR Challenge}

The Student-Code Optical Character Recognition (OCR) challenge is to convert images of handwritten student code into a digital string representation. This process presents a significant challenge, particularly for languages like Python where indentation plays a crucial semantic role. The advent of Large Language Models (LLMs) for post-processing has ushered in an exciting era for OCR technology, enhancing its effectiveness and accuracy. However, the transcription of student code introduces distinct challenges. It is imperative that the OCR process avoids introducing or "hallucinating" logical corrections or fixes. This is true regardless of whether the OCR is conducted by teachers for assessment purposes or by students during their learning process. 

We contribute two novel benchmark datasets as well as evaluation methodology. The \textit{Correct Student Dataset} is created by 40 real students in, Code In Place\cite{piech2021code}, an international, free, online intro to Python course. The students wrote answers to a provided task. The Python coding tasks included grid-world, console, and graphics challenges from the introductory computer science course CS-106A, at Stanford. We augment the dataset with the \textit{Logical Error Dataset} that has a range of mistakes that you would expect to see in an introduction to coding class. While some of the images in the \textit{Logical Error Dataset} were from real students, the majority were written by the authors. We open-source both datasets. Using these two datasets we can then run both the Edit Distance Test to measure how accurately a student-code OCR algorithm captures what the student had written as well as the Logical Fix Hallucination Test to measure how many corrections the OCR method injects.

\begin{table}[t]
\begin{tabular}{@{}lccc}
     \toprule
        \textbf{Dataset}
        & \textbf{Num Photos} 
        & \textbf{Correct}  
        & \textbf{Annotated}
        \\
    \midrule
    Correct Student Dataset & 44 & 100\% & \checkmark \\
    Logical Error Dataset & 11 & 0\% & \checkmark \\
    \bottomrule
  \end{tabular}
  \caption{Two evaluation datasets that we open source. The correct column is the \% of programs with no logical errors. All programs are annotated with what the OCR should produce.}
  \label{tab:datasets}
\end{table}


\subsection{Measuring OCR Error}

Understanding how closely the digitalization from a Student-Code OCR algorithm approximates the true gold label digitalization is crucial. For the 55 programs in both the \textit{Correct Student Dataset} and the \textit{Logical Error Dataset} combined, we measure the average Levenshtein distance, normalized by the length of the student programs. The Levenshtein distance quantifies the minimal number of single-character edits (i.e., insertions, deletions, or substitutions) required to change the predicted student program into the annotation of what they actually meant to write \cite{yujian2007normalized}. 
Lower values of the edit distance indicate that the algorithm's output is closer to what the student intended. By normalizing the Levenshtein distance by the number of characters in the test we have a metric that is more interpretable. It is approximately, the \% of characters for which the algorithm made a mistake. Mathematically, the normalized Levenshtein distance (\( L_{\text{norm}} \)) is defined as follows:

\[ L_{\text{norm}} = \frac{L(str_1, str_2)}{|str_1|} \times 100\% \]

where:
\( L(str_1, str_2) \) is the Levenshtein distance between the ground truth string \( str_1 \) and the OCR output string \( str_2 \).
 \( |str_1| \) denotes the length of the ground truth string. The OCR Error metric is the average $L_{\text{norm}}$ across all student handwritten programs.

\subsection{Logical Fix Hallucination Test}

One of the primary concerns in integrating AI systems that possess vast knowledge of coding is the potential risk of the system inadvertently providing answers to the students. Take, for example, a scenario wherein a student is attempting to write a solution to the Fibonacci sequence, a rather standard programming task. An AI system, especially one equipped with a language model, might unintentionally rectify conceptual errors in the student's code, leading to what we term as "logical fix hallucination". This not only risks undermining the student's learning process but also doesn't represent a good faith translation of the student's original intention. To safeguard against this and to accurately measure the extent of solution hallucination, a robust testing method is essential. In response to this need, we introduced the \textit{Logical Error Dataset}. This dataset comprises a curated selection of typical student errors, thereby serving as a benchmark to determine whether the AI recognition algorithm tends to hallucinate fixes.

The \textit{Logical Error Dataset} includes 11 programs. Each program has an image of the program handwritten, the correct digitalization, and the error description (see Table~\ref{tab:logical_errors} for description). The dataset contains a spectrum of errors such as Fence Post Errors,  Arithmetic Errors, Control Flow Errors, and Scope Errors. These errors focus on logical, or semantic, errors rather than syntactic errors (such as misspelling a variable or forgetting a colon). This is for a pedagogical reason: syntactic issues are often ones that a teacher may teach by showing the correction. Logical errors are ones where the teacher may want the student to find the solution on their own.


\section{Methods}


We consider two methods for OCR of student code: An algorithm that uses LLMs as post-processing as well as a multimodal LLM. In the former algorithm, we decompose the task into three phases. (1) In the first phase we apply an off-the-shelf OCR method, such as Azure or Google OCR. (2) In the second phase we apply an Indentation Recognition algorithm and (3) in the third and final phase we use an LLM for Post Correction.

\subsection{Initial OCR}

The initial phase involved digitizing the handwritten code, for which we employed four leading Optical Character Recognition (OCR) technologies: Google Cloud Vision, Microsoft Azure OCR, AWS Textract, and MathPix. These platforms established our baseline for accuracy assessment. As demonstrated in \hyperlink{tab:results1}{Table 2}, the error rates were significantly high, underscoring the challenges of OCR in the given context. Minor typographical errors, inherent to the OCR process, rendered the codes non-executable, highlighting the critical need for precise transcription in coding applications.

This observation prompted further investigation into two key areas: recognition of indentation patterns, essential for understanding code structure, and the development of a post-correction mechanism to rectify OCR-induced errors. Given the comparative analysis, Microsoft Azure OCR was selected for subsequent phases due to its superior accuracy among the evaluated platforms. 

\subsection{Indentation Recognition}

\begin{figure*}[ht]
    \centering
    \includegraphics[width=\linewidth]{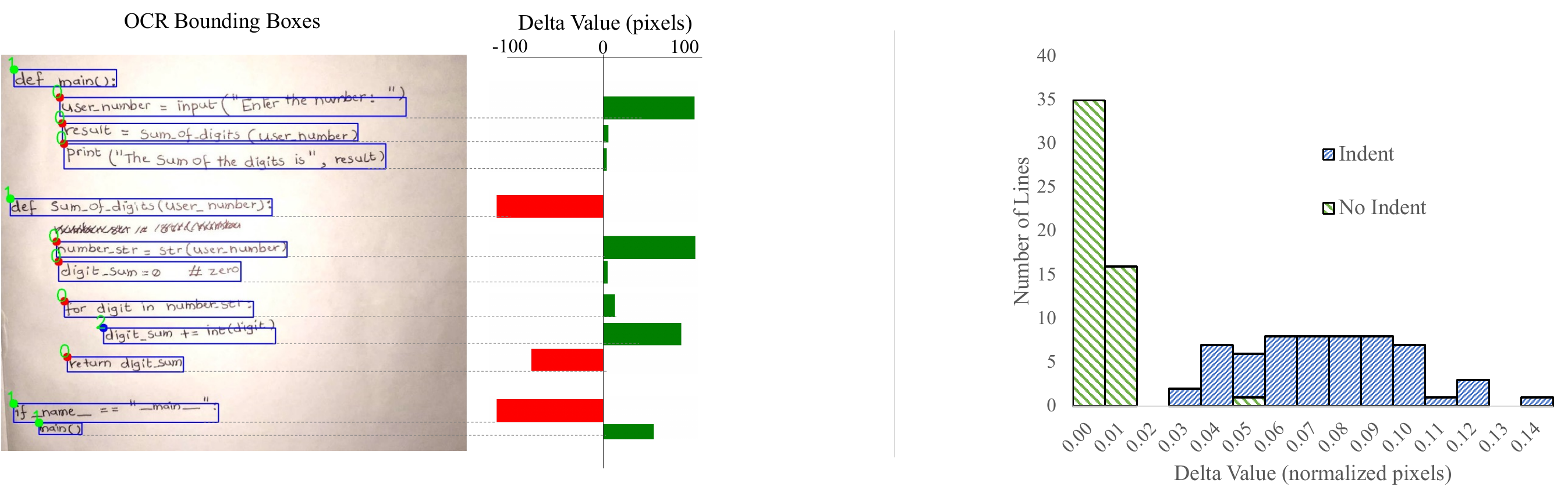} 
    \caption{Left: Example program with corresponding delta values. Large positive deltas signify indentation. Right: A histogram of positive delta values among 16 images shows a clear distinction between indent and no-indent.}
    \label{fig:bootstrap-histogram}
\end{figure*}

Indentation is a critical part of OCR for handwritten Python code. In Python indentation has semantic meaning. However, raw OCR results do not preserve the indentation structure intrinsic to the ground truth. The OCR data does includes bounding box coordinates for each line of text, providing spatial information that could be leveraged to deduce indentation levels. We utilized this spatial data through two different methods to ascertain the indentation level for each line of code.

\paragraph{\textbf{Absolute Indentation Clustering}}
Indentation recognition in handwritten code can be formalized as a clustering challenge, predicated on the assumption that lines with similar indentations would align at comparable horizontal start points. The main task is to cluster the x-coordinate of the top-left point of each line's bounding box, as identified by OCR, to identify distinct indentation levels. This effectively transforms the problem into a straightforward one-dimensional clustering task. We used Meanshift clustering as it does not require a predefined number of clusters \cite{cheng1995mean, comaniciu2002mean}

Meanshift has a bandwidth hyperparameter, which defines the range of influence of each cluster centroid. Across our datasets, no bandwidth worked well for all images. To address this variability, we devised an adaptive bandwidth estimation formula: $
\text{Estimated Bandwidth} = 1.5 \times \frac{1}{N} \sum_{i=1}^{N} h_i
$, where \( h_i \) represents the height of the \( i^{th} \) bounding box in the OCR output, and \( N \) is the total number of bounding boxes.





\paragraph{\textbf{Relative Indentation Clustering}}

We hypothesize that when students write code, the horizontal spacing of each line is influenced by the spatial position of the immediately preceding lines. We propose a method that uses the relative difference between lines. 

In our relative indentation approach, we translate the OCR bounding box outputs into "deltas" between each line's minimum "x" coordinate (See Figure 2, left side). So, for an image with \( n \) lines of code, we will have \( n-1 \) deltas. We normalize the deltas using the image's width.  We observe from visual inspection that using these deltas, it is much easier to differentiate between indentation and no indentation. See Figure 2. If we consolidate the deltas with positive values we notice that some positive delta values are large, corresponding to an indentation. Others are close to zero, corresponding to no indent. 



The relative indentation method separates deltas into ones with positive and negative values.  Among the lines with positive indentation, we classify the deltas as either being "single indent" or "no indent". Among lines with a negative delta, we search for the nearest ancestor. See Algorithm 1 for details.
\begin{algorithm}[h!]
\caption{Relative Indentation Algorithm} 
For each line \( L_i \), starting from the second line (\( i = 2 \)) to the last line (\( i = n \)), the delta \( \delta_i \) between its minimum x-coordinate and that of the preceding line \( L_{i-1} \) is examined:
\begin{enumerate}
    \item \textbf{Positive Delta}: If \( \delta_i > 0 \), the line is evaluated for potential indentation changes. If \( \text{Cluster}(\delta_i) = \text{Indent} \), the indentation level of \( L_i \) is incremented by one. If \( \text{Cluster}(\delta_i) = \text{No-Indent} \), the indentation level of \( L_i \) remains unchanged.
    
    \item \textbf{Negative Delta}: If \( \delta_i < 0 \), the "Nearest Ancestor" technique is applied:
    \begin{enumerate}
        \item Form a dictionary \( D \) by tracing upwards from line \( L_i \), recording the first line \( L_j \) for each unique indentation level encountered, until the beginning of the code is reached.
        
        \item The indentation level for \( L_i \) is determined based on the smallest absolute difference \( |\delta_{i,j}| \) between \( L_i \)'s minimum x-coordinate and those of lines \( L_j \) in \( D \), indicating the closest horizontal alignment. Specifically, \( L_i \) is aligned with the indentation level of \( L_j \) in \( D \) that minimizes \( |\delta_{i,j}| \).
    \end{enumerate}
\end{enumerate}
\end{algorithm}

To classify between indent and no-indent, we model the positive deltas using a Gaussian Mixture Model (GMM). The GMM has two Gaussians, one to represent the deltas for Indent and one to represent deltas for No-Indent:
\begin{align*}
D_{\text{no-indent}} &\sim N(\mu_1, \sigma_1^2)\\
D_{\text{indent}} &\sim N(\mu_2, \sigma_2^2)
\end{align*}

Our apriori assumption is that each Gaussian is equally likely. Formally, that is equivalent to setting the mixture parameter $\tau =0.5$. As such, to train our model we simply need to estimate the mean ($\mu$) and variance ($\sigma$) of the two Gaussians. We estimate the four parameters using LOOCV on a subset of manually annotated images from our data. We fit the hyper-parameters of the GMM using Maximum Likelihood Estimation (MLE) from the labelled data. This  produces the following estimates for the hyper-parameters:
\begin{table}[ht]
\centering
\begin{tabular}{lcc}
\hline
            & \textbf{Mean}        & \textbf{Standard Deviation} \\ \midrule
\textbf{No-Indent}    & $\mu_1$ = 0.007    & $\sigma_1$ =0.008           \\ 
\textbf{Indent}  & $\mu_2$ = 0.078    &  $\sigma_2$ =0.025           \\ \hline
\end{tabular}
\label{tab:mean-std}
\end{table}

We can now classify if a delta pixels $\delta$ is an indentation using Bayes' Theorem. Let $f$ be the normal probability density function: 
\begin{align*}
P(\text{Cluster}(\delta) = \text{Indent}
) &= \frac{f(D_{\text{indent}} = \delta)}{f(D_{\text{indent}} = \delta)+ f(D_{\text{no-indent}} = \delta) } 
\end{align*}

See Figure 2 (right side) for the different delta values between Indent and No-Indent lines of code among the 16 images. We observe that there is a clear split among the two groups, which suggests that the classification would be robust to different hyperparameter values. We note that it would be slightly more accurate to model $D_{\text{no-indent}}$ as a truncated normal (given that its value can never be less than 0). However, given that the groups Indent and No-Indent are so well separated, the model is robust to using a classic GMM and having no prior on indentation.

\subsection{Language Model-Based Post-Correction}

The tiniest of textual errors introduced by OCR can render a code un-executable. As the last part of our method, we use a language model to fix the errors in transcription introduced by the OCR. This module receives input directly from the Indentation Recognition Module, with the explicit aim of correcting only the typographical errors without altering the code's logical structure or indentation. A significant challenge in this phase is mitigating the "hallucination" effect commonly observed in large language models---unintended alterations such as indentation changes, unwarranted corrections of logical errors, or random modifications. Our objective was to minimize the edit distance, ensuring no deviation from the original text. We explored two distinct approaches.\footnote{The exact prompts used for the experiment can be found in the \href{https://github.com/mdoumbouya/codeocr/blob/main/src/code_ocr/post_correction.py}{\textbf{post-correction module in the codebase}}. Specifically, we used \textit{SIMPLEprompting\_test2}, \textit{COTprompting\_test5}, and \textit{GPT4\_Vision} for reporting our results.}

\paragraph{Simple Prompting Approach}

The initial strategy involved straightforwardly feeding the output from the Indentation Recognition Module to the language model. This approach significantly reduces errors but is susceptible to a hallucination-based logical fix. The challenge was engineering a prompt that maintained a low error rate while minimizing hallucinations. We found it particularly important to include direct messaging such as: "*VERY STRICT RULE* - Do not fix any logical, or numerical error of the original code. - Do not fix any indentation of the original code."






\paragraph{Chain-of-Thought (CoT) Prompting Approach}

\begin{figure}[h!t]
    \centering
    \includegraphics[width=\linewidth]{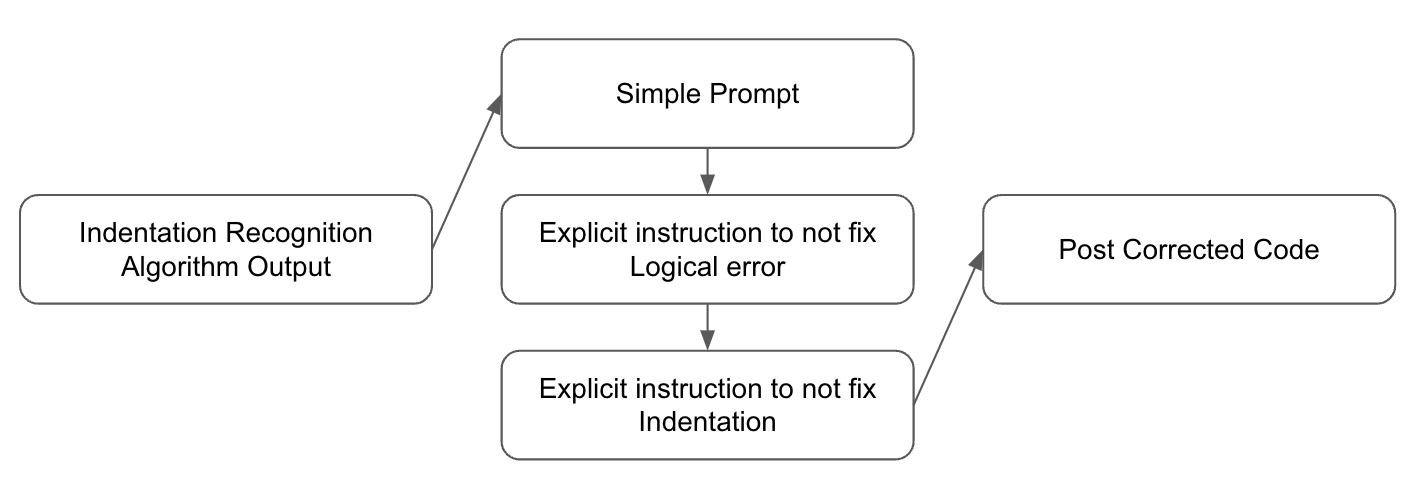}
    \caption{Chain-of-Thought Prompting Pipeline}
    \label{fig:cot-prompt-pipeline}
\end{figure}

To further reduce hallucination, especially regarding logical corrections and indentation changes, we adopted a more nuanced, three-step Chain-of-Thought Prompting Approach. The first step involves prompting the language model to correct spelling errors potentially introduced by the OCR system. Subsequent steps assume that the model might inadvertently correct logical errors or alter indentation; hence, specific instructions are provided to revert any such changes. This method proved highly effective and eliminated all hallucinatory logical corrections in our dataset. However, we observed a slightly higher OCR error rate in this case compared to Simple Prompting. The Chain-of-Thought Prompting Pipeline is depicted in Figure \ref{fig:cot-prompt-pipeline}

\subsection{Multi-Modal Handwritten OCR}
Large multi-modal models offer robust text transcription directly from images, enabling an end-to-end transcription process that mitigates the need for intermediate steps like our three-stage approach. Leveraging GPT-4-Vision-Preview, we implemented an end-to-end OCR process that directly transcribes text from images without the need for segmented preprocessing steps like the aforementioned method. This method involves directly feeding images of handwritten code, with a prompt into the model. The model then applies its capabilities to recognize and transcribe text. The key to this approach is a carefully crafted prompt that guides the model to focus strictly on transcription while avoiding the introduction of errors typical of automated recognition systems. The specific prompt used, instructs the model to strictly adhere to the text as it appears in the image, emphasizing accuracy and fidelity to the source without attempting to correct or interpret the code logic or structure.

\section{Results}

\begin{table}[t!]
\begin{tabular}{@{}lll}
     \toprule
     
        & \textbf{OCR Error} & \textbf{Logical Fix}  \\
    \midrule
    \textbf{OCR Algorithm} \\
    \hspace{1mm}
    Azure & $30.2 \pm 1.8$\% & 0\% \\
    \hspace{1mm}
    Google & $39.4 \pm 2.3$\% & 0\% \\
    \hspace{1mm}
    AWS & $34.4 \pm 2.3$\% & 0\% \\
    \hspace{1mm}
    MathPix & $64.6 \pm 5.5$\% & 0\% \\
    \hspace{1mm}
    GPT 4V & $6.0 \pm 0.8$\% & 5 $\pm$ 1\% \\

    \textbf{Indentation Recognition} \\
    \hspace{1mm}
    Azure + No  Fix & $30.2 \pm 1.8$\% & 0\% \\
    \hspace{1mm}
    Azure + Absolute & $28.3 \pm 5.0$\% & 0\% \\
    \hspace{1mm}
    Azure + Relative & $20.2 \pm 2.4$\% & 0\% \\

    \textbf{Post Correction} \\
    \hspace{1mm}
    Azure + Relative + No Post & $20.2 \pm 2.4$\% & 0\% \\
    \hspace{1mm}
    Azure + Relative + CoT & $8.5 \pm 1.0$\% & \textbf{0\%} \\
    \hspace{1mm}
    Azure + Relative + Simple & \textbf{5.3 $\pm$ 0.9\%} & 9 $\pm$ 2\% \\
    \bottomrule
  \end{tabular}
  \captionof{table}{Main Results: In the top section we compare off the shelf OCR algorithms. In the middle section, we show the benefits of including indentation recognition. In the bottom section, we show the improvement from applying post-correction methods. OCR Error is the average normalized Levenshtein distance across all 55 programs (see section 2.1 for details). Logical Fixes are the percentage of the 11 programs where the algorithm injects a semantic correction to student code. $\pm$ is standard error.}
  \label{tab:results1}
\end{table}


The off-the-shelf, state-of-the-art algorithms performed quite poorly on the OCR challenge for handwritten student code. Among the commercial solutions, Azure was the best performing with an average OCR Error (normalized Levenshtein distance) of $30.2 \pm 1.8$.  As an aside, while MathPix had a high error rate, that was mainly due to its representation of code as LaTeX. It was still useful when combined with LLM prompting, though not as accurate as Azure.

By employing our algorithms to correct indentation and by using LLM post-correction we were able to decrease OCR error significantly. The best-performing algorithms used our GMM-based relative indentation correction and LLM post-correction. The two methods of post-correction had different advantages. The "Simple" method achieved the lowest OCR error ($5.3 \pm 0.9$). While the Chain of Thought post-correction technique had a higher OCR error ($8.5 \pm 1.0$) it produced 0 logical fixes, compared "Simple" which fixed 9\% of the errors in the \textit{Logical Error Dataset}. See Table~\ref{tab:results1}.



\textbf{Multi Modal Results:} The large Multi-Modal GPT-4V(ision) achieved notable success, registering an average OCR Error of $6.0 \pm 0.8$ while only introducing logical fixes in 5\% of the \textit{Logical Error Dataset}. These results are promising and appear to approach the Pareto frontier between the two post-correction methods.

 \textbf{Indentation Algorithm Results:} Both the absolute algorithm (Mean Shift) and the relative algorithm (GMM-based) improved OCR error rates. The relative indentation algorithm had the best results and qualitatively seemed to make very few indentation errors. This was especially important for "Grid World" programs \cite{becker2001teaching} where the LLM would not have been able to infer the indentation level simply from the code. The Mean Shift was accurate but would make mistakes, especially on longer programs. Lines that, to a human observer, would belong to one indentation level might be situated closer on the x-axis to a denser cluster associated with a different indentation level. Such discrepancies highlighted the limitations of absolute indentation clustering in contexts where local data characteristics may offer a more intuitive guide to cluster membership than global data density. The four hyper-parameters for the Relative indentation algorithm were set using a subset of 16 images. Even though there are only four hyper-parameters, those values might have overfit the indentation statistics of those 16 images. To make sure that the indentation results are valid, we rerun all of the results on the 39 images that were not used to set the hyper-parameters. In this "heldout set" we observe that Relative indentation recognition has a similar success as on the full dataset. It performs just as well both before and after the post-correction (See Table \ref{tab:filtered-results1} in the appendix for the full results). This gives us confidence that the four hyper-parameters did not allow the relative indentation algorithm to overfit the data.

\subsection{Qualitative Analysis of Language Model Hallucinations, and Logical Fixes}

The prompting techniques that we used did not induce substantial hallucinations of logical fixes. The best-performing algorithm in terms of OCR error (Azure + Relative + Simple) had a single non-indentation logical fix out of the 11 programs in the\textit{ Logical Error Dataset}. This fix can be seen in Figure \ref{fig:example-52}.

\begin{figure*}[t]
    \centering
    \includegraphics[width=\textwidth, trim=0 9cm 0 0, clip]{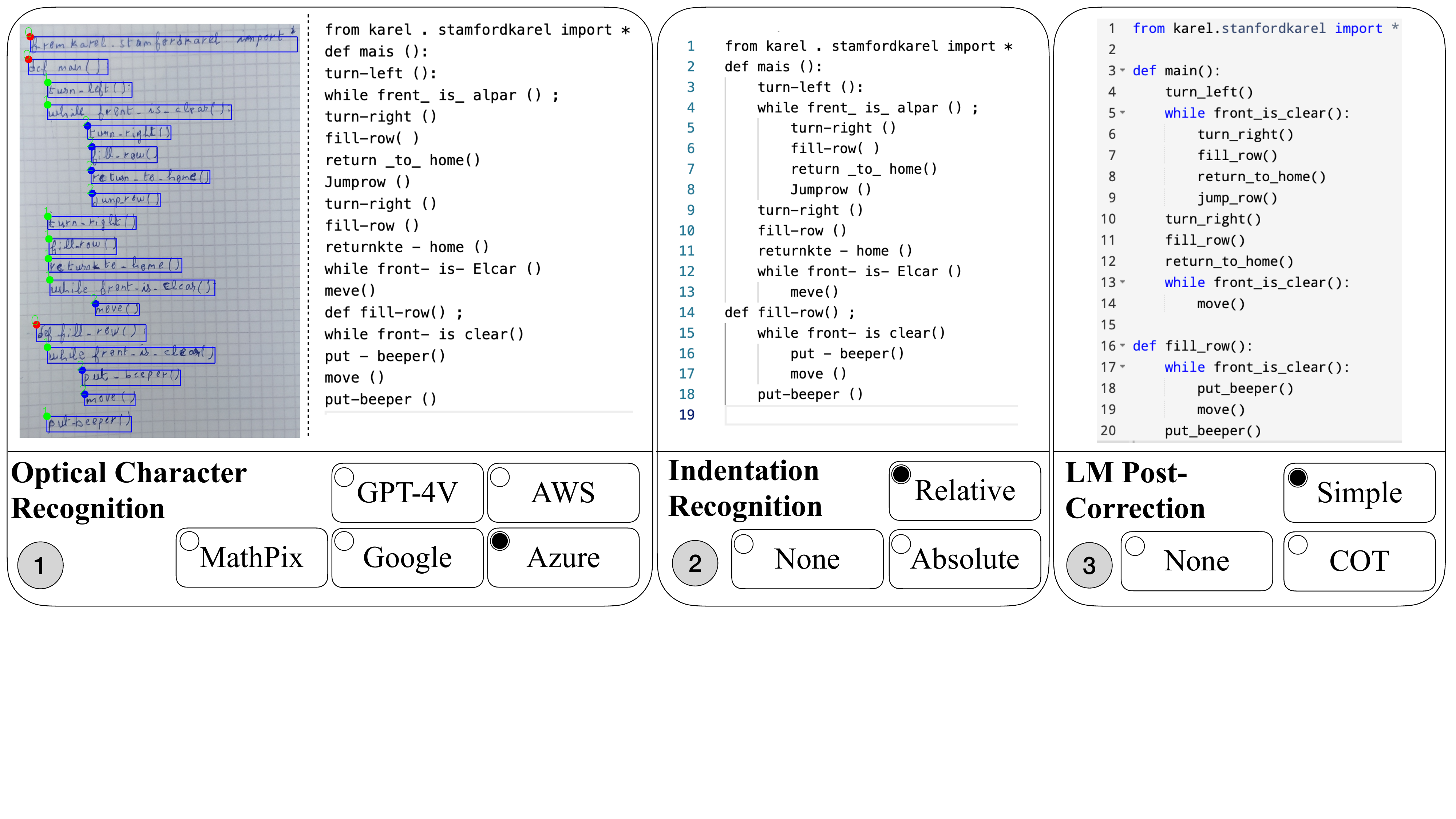}
    \caption{Example of perfect transcription of a student's program in a grid world programming environment which is often used in introductory programming courses.
    The language model accurately fixed all OCR errors and did not introduce any artifacts, despite this type of program being less frequent in the training set of large language models.
    }
    \label{fig:grid-world-programming}
\end{figure*}

\begin{figure*}[t]
    \centering
    \includegraphics[width=\linewidth]{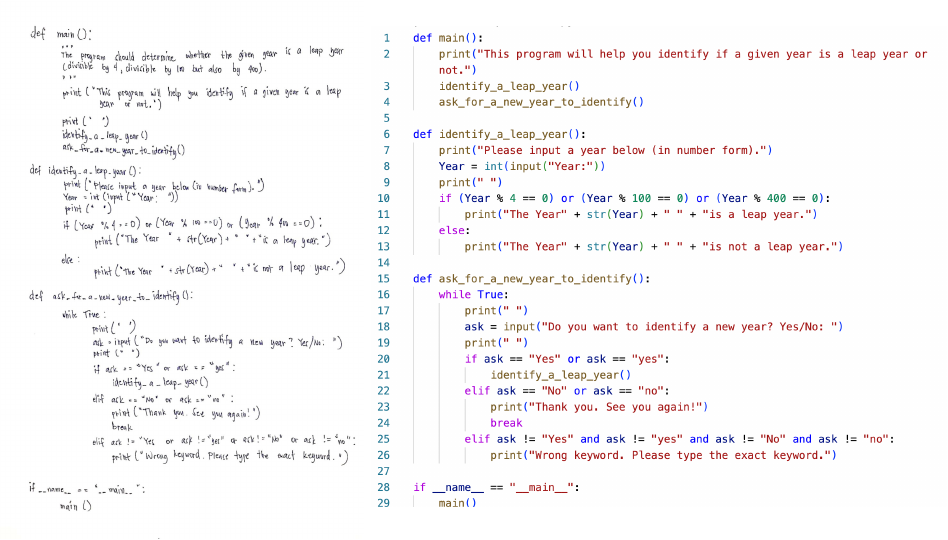}
    \caption{An example of OCR (Azure + Relative + Simple) on a longer example of handwritten student code. The OCR faithfully translates the student program, even keeping logical errors such as the test for leap year. It does not keep the student's comment and it does not include the second "print" line: \texttt{print(" ")}}
    \label{fig:long-example}
\end{figure*}

On line 4 the student wrote \texttt{if number / 2 != 0} but the LLM corrected it to be \texttt{if number \% 2 != 0}. Note that the algorithm changed the incorrect division to be a mod -- possibly giving away part of the answer to the student. For this particular example, the Chain of Thought prompting did not create the logical fix, however, it incorrectly translated the division as a 1 (carrying forward a mistake from Azure). For some learning environments, this type of logical fix may be more tolerable than others. There are many examples where the simple OCR algorithm managed to faithfully translate the student's code while still maintaining the logical errors. One example can be seen in Figure \ref{fig:long-example}. In the identify leap year function the student included the incorrect logic \texttt{if (Year \% 4 == 0) or (Year \% 100 == 0) or (year \% 400 == 0)}. It should have been \texttt{if (Year \% 4 == 0) and (Year \% 100 != 0) or (Year \% 400 == 0)}. Even though the LLM certainly would be able to produce the correct code, it did not fix the student's mistake.

\section{Limitations}
\subsection{More comprehensive evaluation datasets}

One of the contributions of this paper is the first (to the best of our knowledge) public datasets of handwritten student codes. However, we hope that for future work we can substantially increase the size of our public dataset. 55 programs is enough to understand high-level differences between algorithms, now that future work will concentrate on more subtle improvements (from 5\% error rate towards 0\%) it will be important to increase the size of this dataset. Similarly, it will be important to fix the \textit{Logical Error Datase}t to encompass a broader set of the sorts of logical errors that students could make while programming. It is surprisingly hard to get handwritten student code which is free to share. Now that we have a useful solution, we hope to be able to collect orders of magnitude more data (with student consent, of course).

\subsection{The Iterative Editing Challenge}

In this paper, we focused on the context of having a single photo that needs to be digitized into code. However, in a natural learning environment, we imagine that students will also need a user interface that supports iterative work. 
A potential interaction with our OCR system could unfold as follows: a student writes code by hand, digitizes it for execution, and corrects a syntax error on the computer, but then needs to revisit the handwritten code for conceptual adjustments. This process raises a question: does the student rewrite the entire code? There's an undeniable need for a more sophisticated approach to such iterative editing. One default solution would be for students to maintain significant whitespace between lines, facilitating subsequent code insertions. Another solution is for students to take photos of different subsets of their code. If they need to edit a part of their code, they would only need to replace the corresponding photo. 

\subsection{Handling Crossed-Out Code}

As shown in the example in Figure 1, our system is able to handle basic "crossing out" of code. We have observed that our system handles crossed-out codes with notable accuracy. However, crossed-out code can become arbitrarily hard. Students writing code, without concern for the OCR system could use annotations that our OCR system is not able to handle. One example of this would be the use of arrows to indicate that a block of code should be inserted somewhere in the codebase. One potential solution is to set expectations for students that they need to keep their code as clean as possible. However, we note that there is great promise that the multi-modal systems, such as GPT4V may be especially adept at handling these sorts of annotations.

\subsection{Applicability to Larger Programs}

The longest programs that we tested in this paper were on the order of 40 lines long. The scalability of our approach to longer programs remains a topic of inquiry. The threshold beyond which our method might be less effective or feasible for learners is yet to be established. For longer programs it seems reasonable to have the student take several photos of their code. While this approach complicates indentation recognition, achieving consistent indentation recognition across multiple photos appears to be a solvable issue.


\subsection{Offline Mode}
Our system has three distinct modules, which could be implemented in various ways: (1) OCR module, (2) Indentation Recognition Module, and (3) Language model used for post-correction.
In the offline embodiment of our system, which we will explore in future work, the OCR module and the language model used for post-correction are executed on a local device, reducing the usage cost (see Appendix \ref{sec:cost}), and removing the need for internet connectivity.





\section{Discussion}
Our research into the digitization of handwritten code, utilizing a symbiotic approach that combines Optical Character Recognition (OCR) with Large Language Models (LLMs), has culminated in both noteworthy outcomes and a comprehensive understanding of the prevailing challenges in this domain. This initiative primarily aimed to augment computer science education in regions where computer accessibility is restricted.

\subsection{CS Education with Limited Digital Resources and Minimal Distractions}
Handwritten code recognition facilitates computer science education without the need for prolonged access to computers, which is particularly beneficial in two educational settings.
In the first setting, students do not have personal computers, and their schools lack the funding to acquire and maintain computer labs. With our solution, a classroom could perform programming activities using only a single shared mobile device to scan and execute handwritten student code.
In the second setting, to mitigate the adverse effects of prolonged screen exposure, teachers may intentionally limit young students' access to computers. Instead, they encourage activities performed on paper with pens. In this scenario, as in the first, students can meticulously craft their programs on paper and then use the classroom's shared code execution device. Handwritten code recognition thus provides a familiar, distraction-free way for students to engage with algorithms and programming.
Additionally, this method fosters reflection, review, and precision, thereby promoting deliberate learning. These experiences instill qualities beyond mere coding skills. By embracing accessible methodologies, educators worldwide can prioritize comprehension and problem-solving skills, empowering students to achieve a deeper understanding of computer science concepts.

\subsection{Application in Grading Handwritten Exams}

In the current academic landscape, digital exams are susceptible to misconduct, especially with the advent of sophisticated LLMs like ChatGPT. Many educational institutions, from high schools to universities, resort to handwritten exams. These exams, while reducing cheating opportunities, are labor-intensive to grade. Additionally, the potential introduction of gender bias during grading—stemming from handwriting perception—is a concern. By digitizing and accurately recognizing text from these handwritten exams, automated unit tests can be employed. This not only expedites the grading process but also may diminish human-induced grading variance. However, to maintain academic integrity, it's paramount that the digitization process avoids any form of solution hallucination.



\subsection{OCR of Grid Based Coding}

In an early stage of our research, we considered whether we could develop a method that would make zero errors. One creative solution was to have students write their code on graph paper in a grid-based representation of their code.
Grid-based coding has been proposed for accessible CS education in the context of low vision and blind programmers \cite{ehtesham2022grid}. OCR of a handwritten grid should be substantially easier than handwritten code. Clearly, this introduces an extra level of work for the student. However, we hypothesized that this extra work could have deep pedagogical benefits. Learning to represent one's code in a binary representation would expose students to some valuable lessons. However, as we realized that OCR for students directly writing Python would be so accurate we did not fully explore this interesting direction.


\section{Conclusion}

We believe that the ability to digitize handwritten student code could have transformative potential for learning coding at scale. It will be especially useful for increasing accessibility to students who don't have or want constant access to a computer. In this paper, we introduce a novel methodology for indentation recognition as well as the first application of LLMs to the task of student code OCR. We contribute two novel datasets for student-code digitalization labelled with the indented code, and including deliberate errors. The tool we have developed is accurate enough to be used by students in a real learning environment. We plan to deploy this research in classrooms in Guinea where the project originated and Bangladesh, as well as in a massive online coding class.


\appendix
\clearpage

\section{Benchmarks}

This is the result when we filtered the 16 images we trained it on.
\label{benchmarks:filtered-resutls1}
\begin{table}[h]
  \centering
  \begin{tabular}{@{}ll}
    \toprule
    \textbf{Method} & \textbf{OCR Error} \\
    \midrule
    \textbf{OCR Algorithm} & \\
    \quad Azure & $32.1 \pm 2.4$ \\
    \quad GPT 4V & $5.9 \pm 1$ \\
    \midrule
    \textbf{Indentation Recognition} & \\
    \quad Azure + No Fix & $32.1 \pm 2.4$ \\
    \quad Azure + Absolute & $32.2 \pm 7$ \\
    \quad Azure + Relative & $23.1 \pm 3.2$ \\
    \midrule
    \textbf{Post Correction} & \\
    \quad Azure + Relative + No Post & $19.85 \pm 1.8$ \\
    \quad Azure + Relative + Chain of Thought & $9.3 \pm 1.3$ \\
    \quad Azure + Relative + Simple & \textbf{5.6} $\pm$ \textbf{1} \\
    \bottomrule
  \end{tabular}
  \caption{OCR Error rates after various correction stages (n=39)}
  \label{tab:filtered-results1}
  \vspace{-2em}

\end{table}

\section{Cost Calculation Breakdown}
\label{sec:cost}
This section provides a breakdown of the cost calculations for the GPT-4(Vision) and our Azure + Relative Indentation + Simple.

The cost analysis is based on the February 2024 pricing for the respective services:

\begin{itemize}
    \item Azure OCR: \$0.001 per image
    \item GPT-4-0613 Text Processing:
        \begin{itemize}
            \item Input: \$0.03 per 1,000 tokens
            \item Output: \$0.06 per 1,000 tokens
        \end{itemize}
    \item GPT-4-Vision-Preview: 
        \begin{itemize}
            \item Image: \$0.00765 (over 768px by 768px)
            \item Input: \$0.01 per 1,000 tokens
            \item Output: \$0.03 per 1,000 tokens
        \end{itemize}
\end{itemize}

Our pipeline integrates Azure OCR and GPT-4 text processing, with the costs broken down as follows:

\begin{itemize}
    \item Azure OCR Cost: \$0.001 per image
    \item GPT-4 Text Processing Cost: Calculated based on the average token count per image
\end{itemize}

In our dataset, the handwritten codes had an average of 320.2545 characters, which roughly equates to 80.0636 tokens (assuming an average of 4 characters per token). 

The instructions comprised 381 characters, contributing to nearly 95.25 tokens. In contrast, the output codes held an average of 341.5455 characters, resulting in around 85.3863 tokens.

The cost for text processing per image is thus calculated as:
\[
\text{Price} = \frac{(80.0636+95.25) \times 0.03 + 85.3863 \times 0.06}{1000} \\ = \$0.01038 
\]

Summing the costs for Azure OCR and GPT-4 text processing yields our total pipeline cost:

\[
\text{Total Pipeline Cost} = \$0.001 + \$0.01038 \\= \$0.01138 \text{ per image}
\]

The GPT-4(Vision) model incurs a cost of \$0.00765 per image for all images over 786x786 size, which is standalone and does not require additional text processing costs.

The instructional requirement for each image is quantified as 387 characters, which translates to roughly 96.75 tokens at an approximate ratio. The output codes held an average of 308.9636 characters, approximately resulting in around 77.2409 tokens.

The cost per image is thus calculated as:
\[
\text{Price} = \frac{96.75 \times 0.01 + 77.2409 \times 0.03}{1000} + 0.00765= \$0.01094
\]

Comparing the costs:

\begin{itemize}
    \item Our Pipeline Cost: \$0.01138 per image
    \item GPT-4(Vision) Cost: \$0.01094 per image
\end{itemize}

\section{Details of Logical Error Dataset}
\label{sec:logical-error-detail}
\begin{table}[h]
\centering
\begin{tabular}{cp{6cm}}
\toprule
\textbf{Image ID} & \textbf{Description of Error} \\ \midrule
29 & In the `identify leap year` function the student used incorrect logic: \texttt{if (Year \% 4 == 0) or (Year \% 100 == 0) or (Year \% 400 == 0)}. Correct logic: \texttt{if (Year \% 4 == 0) and (Year \% 100 != 0) or (Year \% 400 == 0)}. \\ \hline
45 & Initialized the variable \texttt{max = 0} inside of the for loop, causing it to reset at every iteration. \\ \hline
46 & Off by one error in string indexing: should be \texttt{str[len(str)-i-1]} instead of \texttt{str[len(str)-i]}. \\ \hline
47 & Recursive factorial function lacks a base case and does not handle \texttt{n = 0}. \\ \hline
48 & Fibonacci sequence indexing error: should be \texttt{sequence[i] + sequence[i+1]}, not \texttt{sequence[i+1] + sequence[i+2]}. Incorrect loop condition: should be \texttt{while len(sequence) < n} instead of \texttt{<= n}. \\ \hline
49 & KeyError raised if \texttt{freq[item]} is called and \texttt{item} is not in dictionary. Solution: check if \texttt{item} is in dictionary before incrementing. \\ \hline
50 & The range in the second loop should start from \texttt{i+1} to avoid repeating elements. \\ \hline
51 & Indentation error in assigning \texttt{final\_list} and unnecessary variable \texttt{final\_loop} resetting inside the loop. \\ \hline
52 & Incorrect operator: used division (\texttt{/}) instead of modulo (\texttt{\%}) for checking even or odd. \\ \hline
53 & Non-alphabetic characters are incorrectly counted as lower case due to an overly broad \texttt{else} condition. \\ \hline
54 & Error in multiplication logic: multiplies every number by 0. Initialize \texttt{total} as \texttt{1} for correct operation. \\ \hline
\end{tabular}
\caption{Logical Errors in Handwritten Code Dataset}
\label{tab:logical_errors}
\end{table}

\section{Additional Examples}

\begin{figure*}[h]
    \centering
    \includegraphics[width=0.90\textwidth, trim=0 19cm 0 0, clip]{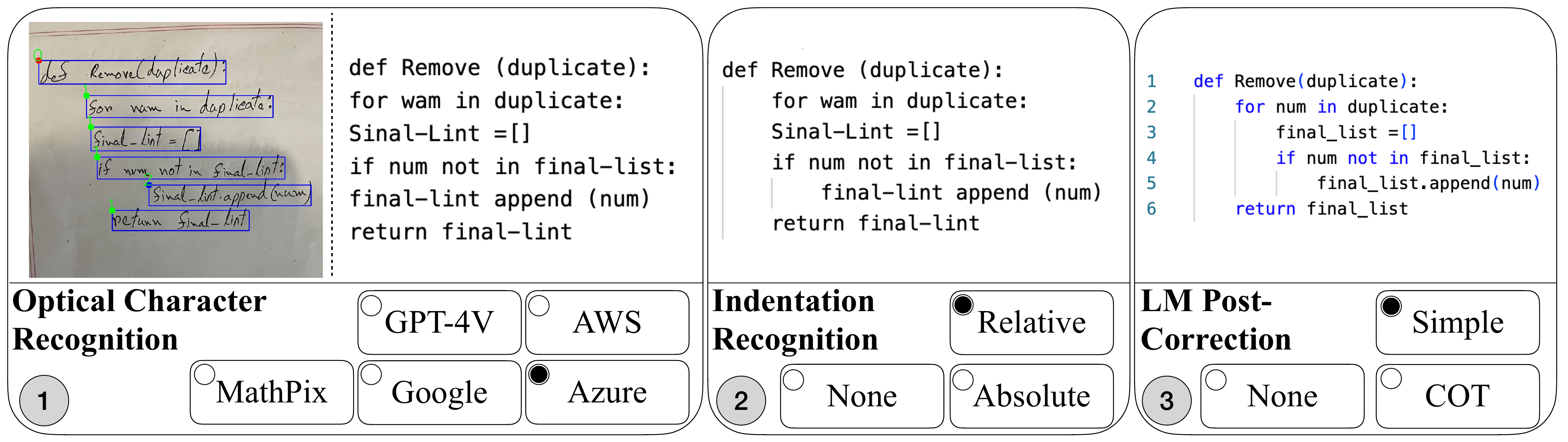}
    
    \includegraphics[width=0.90\textwidth, trim=0 19cm 0 0, clip]{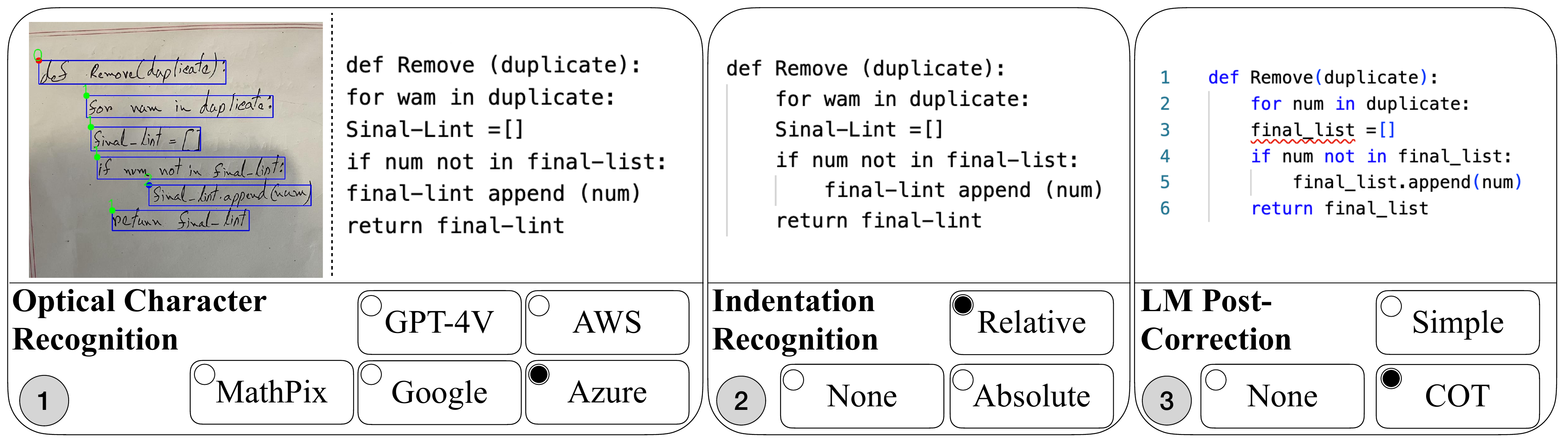}

    \includegraphics[width=0.90\textwidth, trim=0 19cm 0 0, clip]{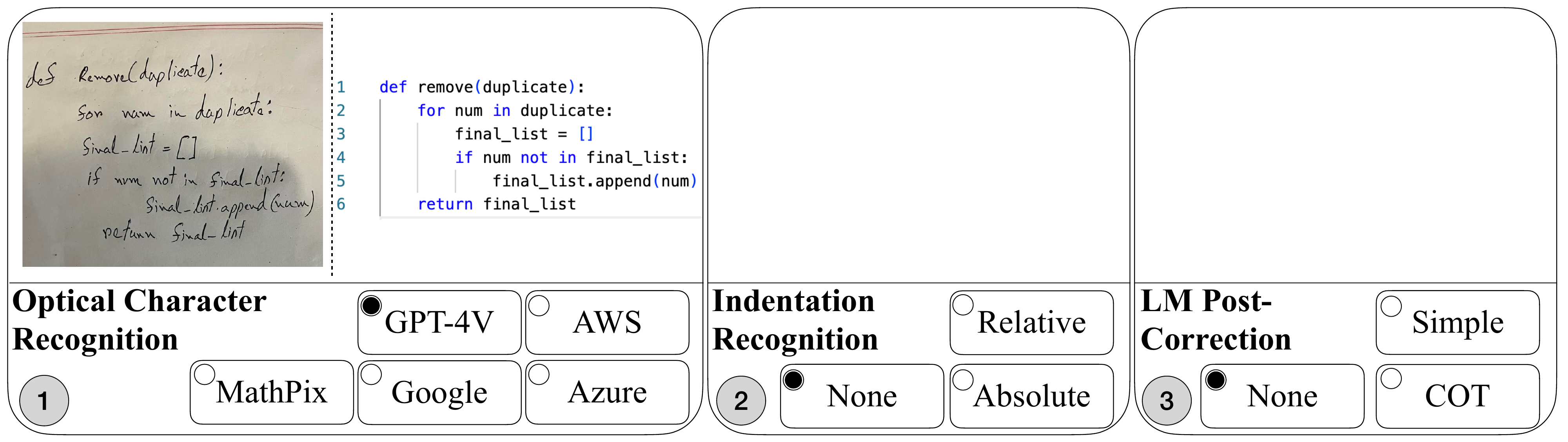}

    \caption{Our system processes a student's handwritten code under various settings. The student made two mistakes: an indentation error, and a variable scoping error (line 3). Top and Bottom: the system hallucinates a fix for the indentation error. Middle: the system correctly transcribed the student's work.}
    \label{fig:example-51}
\end{figure*}

\begin{figure*}[h]
    \centering
    \includegraphics[width=0.90\textwidth, trim=0 19cm 0 0, clip]{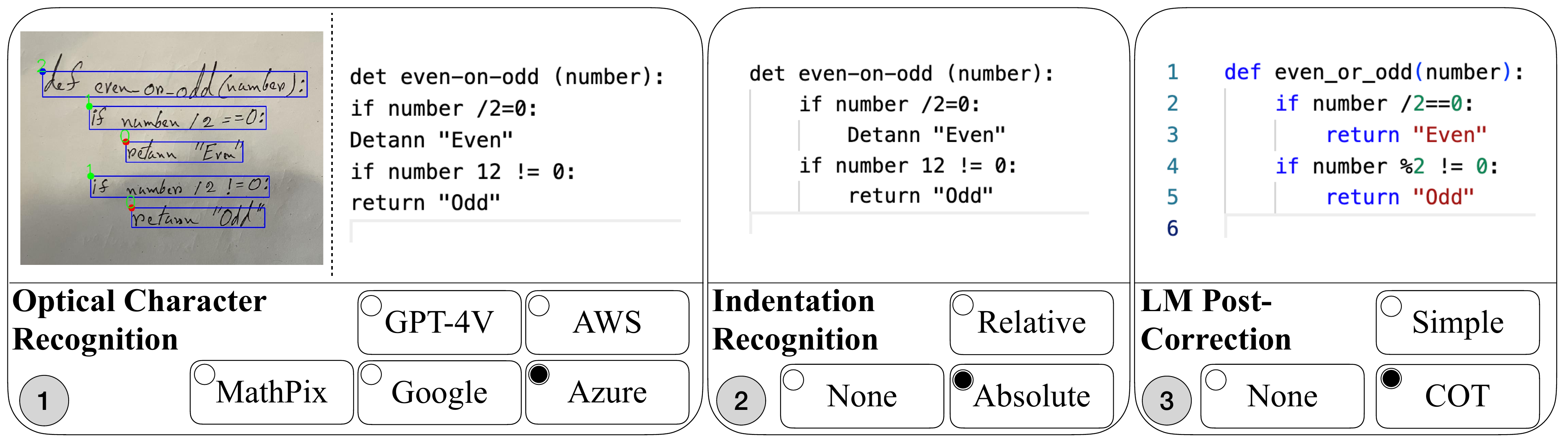}

    \caption{The student made a mistake (used division instead of modulo). The system hallucinated a correction.}
    \label{fig:example-52}
\end{figure*}





\vfill\eject
\bibliographystyle{abbrv}
\balance
\bibliography{main} 


\end{document}